\documentclass[letterpaper]{article} 
\usepackage{aaai2027}  
\usepackage[hyphens]{url}  
\usepackage{graphicx} 
\usepackage{natbib}  
\usepackage{caption} 
\usepackage{algorithm}
\usepackage{algorithmic}
\usepackage{newfloat}
\usepackage{listings}
\usepackage{amsmath}
\usepackage{amssymb}
\DeclareCaptionStyle{ruled}{labelfont=normalfont,labelsep=colon,strut=off} 
\floatstyle{ruled}
\newfloat{listing}{tb}{lst}{}
\floatname{listing}{Listing}
\usepackage{booktabs}
\title{MOSAIC: Addressing Disproportionate Task–Modality Dependencies via Modality-Partitioned Gradient Routing}

\author{
    Aryan Chowdhury,
    Yizheng Sun,
    Thomas Carroll,
    Riza Batista-Navarro
}
\affiliations{
    The University of Manchester
}

\begin{document}

\maketitle

\begin{abstract}
In multi-task settings, combining multiple modalities allows task representations to draw on signals that a single modality alone would not capture. However, individual tasks may depend disproportionately on particular modalities, a structure that conventional fusion-based architectures do not model explicitly, instead enforcing a unified representation space. Consequently, a task’s gradients can affect parameters tied to modalities it barely relies on, leaving features that compromise between competing demands. Although prior work has explored task-weighting and loss-balancing strategies, their effectiveness is limited by the absence of explicit modality separation, preventing models from optimising each modality according to its task-specific importance. To address these limitations, we propose MOSAIC (MOdality-Separated Architecture with Inter-modal Cross-attention), which preserves unimodal representations while enabling controlled interaction between them, and uses modality-specific losses to route gradients to their corresponding partitions. On the Corpus of Everyday Japanese Conversation (CEJC), MOSAIC improves average macro-F1 by 7.4 points across four tasks with a 72\% reduction in backbone parameters relative to a fusion-based baseline. Our results demonstrate that modality partitioning yields consistent gains across tasks while requiring far fewer backbone parameters than a fusion-based design.
\end{abstract}

\section{Introduction}

In multimodal multi-task learning, every task depends differently on each modality. However, existing multimodal transformers predominantly rely on fusion-based architectures, which project modality-specific features into a shared latent space to enable cross-modal interaction \cite{xu2023multimodal, lu2019vilbert, chen2020uniter}. While effective, this unified space forces tasks with different modality needs to compete for the same representational capacity \cite{standley2020tasks}, producing conflicting gradient updates that improve one task at the expense of another \cite{yu2020gradient}. This is compounded by modality interference, where gradients that benefit one modality degrade another \cite{wu2022characterizing}, and modality dominance, where easier modalities overshadow weaker ones, leaving them underused and overfitting at different rates \cite{wang2020makes, huang2022modality}. Forced to balance these competing demands, the shared representation becomes overloaded, yielding reduced task-specific specialisation. The resulting models are also often over-parameterised, modelling cross-modal dependencies that may be unnecessary for any given task \cite{nagrani2021attention, held2023shapley}.

Prior work has largely addressed these problems in isolation. Modality-balancing methods encourage more even modality use by reweighting losses or modulating the gradients of the dominant modality \cite{wang2020makes, peng2022balanced}, but retain a shared representation in which tasks still compete for the same parameters. Multi-task optimisation methods tackle task conflict more directly by rescaling losses or modifying gradients \cite{kendall2018multi, yu2020gradient}, but they are agnostic to modality structure and correct interference after it arises rather than preventing it by design. Modality-specific architectures come closest to a structural solution, assigning separate experts or weights to individual modalities \cite{bao2022vlmo, wang2023image}, yet they keep self-attention global across modalities and do not optimise each modality according to its task-specific importance. To our knowledge, no prior work combines modality-specific parameters with a separate loss for each modality, weighted by task relevance.

In this work, we introduce MOSAIC (MOdality-Separated Architecture with Inter-modal Cross-attention), a multimodal multi-task architecture with modality-specific parameter spaces, each shaped by tasks according to their reliance on that modality. Our contributions are as follows:

\begin{enumerate}

\item A modality-separated transformer architecture that preserves unimodal representations by assigning each modality its own parameter partition while allowing interaction only through cross-attention between partitions at designated layers.

\item Modality-specific losses, defined as a weighted combination of task objectives that reflects each task’s overall importance and its relevance to that modality, with the resulting gradients routed to the corresponding parameter partitions.

\item An evaluation on CEJC across four conversation understanding tasks, achieving a 7.4-point improvement in average macro-F1 while using 72\% fewer backbone parameters than a fusion-based baseline.

\end{enumerate}

By separating what each modality learns and controlling how and where they interact, MOSAIC shows that multimodal multi-task models can be both more accurate and more efficient than fusion-based ones.

\section{Related Work}
\subsection{Multimodal Fusion Architectures}

Multimodal transformers differ mainly in where and how modalities first interact. Early fusion jointly embeds all modalities at the input and processes them with a single shared transformer, which allows dense cross-modal interaction from the first layer \cite{li2019visualbert, chen2020uniter}. However, its cost grows with the combined sequence length, and one stream must handle modalities with very different statistics. Late fusion avoids this by encoding each modality independently and merging only their high-level representations or predictions, missing fine-grained cross-modal dependencies \cite{baltrusaitis2019multimodal}. Dual-stream architectures keep a separate stack per modality, interacting only through cross-attention \cite{lu2019vilbert, tan2019lxmert, tsai2019multimodal}, but the number of stacks scales with the number of modalities, on top of the cross-attention overhead. Bottleneck fusion reduces this cost by routing exchange through a small set of shared latent tokens \cite{nagrani2021attention}, yet the narrow channel limits token-level alignment, and nothing controls how much each modality contributes to those tokens. These designs determine how modalities interact but do not explicitly control how much each should contribute, leaving open whether they contribute equally in practice.

\subsection{Modality Imbalance and Dominance}

In practice, they often do not. Jointly trained multimodal networks can underperform their best unimodal counterparts because modalities overfit and generalise at different rates \cite{wang2020makes}. This is attributed to greedy learning, which fits the easiest modality first and leaves the others under-trained \cite{wu2022characterizing}, and to competition between modalities during joint training \cite{huang2022modality}.

Existing remedies operate at three levels. Gradient-based methods slow the dominant modality by scaling down its gradient according to its contribution relative to the other modalities \cite{peng2022balanced}, or by modulating gradients using each modality's individual contribution \cite{li2023boosting}. Others reconcile unimodal and multimodal gradients through a shared Pareto-based update direction \cite{wei2024mmpareto}. Representation-level methods rebalance learning through class prototypes that speed up the slower modality and restrain the dominant one \cite{fan2023pmr}, or by re-initialising the encoder of the under-learned modality \cite{wei2024diagnosing}. At the data level, modality dropout randomly withholds modalities during training to discourage reliance on any one of them \cite{neverova2015moddrop}. However, these methods are typically evaluated as independent single-task classification problems, largely on audio-visual benchmarks, and do not explicitly model how much each task depends on each modality.

\subsection{Multi-task Learning}

These problems are exacerbated in multi-task settings. Multi-task learning shares parameters across tasks so that related tasks can regularise one another \cite{caruana1997multitask, ruder2017overview}, but whether sharing helps depends on which tasks are grouped \cite{standley2020tasks}.

Loss-weighting methods scale each task's loss by learned uncertainty \cite{kendall2018multi}, by gradient magnitude and learning pace \cite{chen2018gradnorm}, or by dynamic weight averaging \cite{liu2019end}. Because they operate on scalar weights over the shared model, they cannot control how a task uses the individual modalities. Gradient-based methods treat multi-task learning as multi-objective optimisation and seek a common descent direction that leads to Pareto optimal solutions \cite{sener2018multi}, project each task’s gradient away from any task gradient it conflicts with \cite{yu2020gradient}, or choose the update that maximises the worst-case improvement across tasks within a neighbourhood of the average gradient \cite{liu2021conflict}. These methods act on the aggregated gradient of shared parameters and treat the modalities feeding those parameters as interchangeable. Architectural methods learn soft sharing between task networks \cite{misra2016cross}, give each task its own gate over a set of shared experts \cite{ma2018modeling}, or turn the shared layers with the highest gradient conflict into task-specific layers \cite{shi2023recon}. Yet separating parameters by task leaves the modalities within each task competing for the same parameters, and ignores that tasks may rely unevenly on different modalities.

\subsection{Modality-Specific Parameters}

A natural response is to give modalities parameters of their own. MISA separates each modality into a modality-invariant and a modality-specific representation before fusing them \cite{hazarika2020misa}, but this separation is only encouraged by auxiliary similarity, difference and reconstruction losses that need weighting, and the architecture itself does not enforce it. Multiway Transformers build the separation into the architecture by sharing self-attention across modalities while giving each modality its own feed-forward expert \cite{bao2022vlmo, wang2023image}. The allocation is, however, fixed by modality and identical for every task, and since only the feed-forward layers are separated, interference in self-attention remains. Mixture-of-experts models instead use a learned router to send each token to a small set of experts, so that different inputs use different parameters \cite{shazeer2017outrageously, fedus2022switch}. Together, these designs give modalities dedicated parameters, but they allocate that capacity by fixed modality or by learned routing, and they do not explicitly model how important each modality is for each task.

\subsection{Parameter Redundancy in Transformers}

Dedicated capacity need not mean additional parameters. Modern transformers are widely over-parameterised, with substantial redundancy in both attention mechanisms and parameter space \cite{michel2019sixteen, voita2019analyzing, aghajanyan2021intrinsic}. Large fractions of attention heads can be pruned with minimal impact on performance, and intrinsic dimensionality analyses show that task-specific adaptation occurs in a much lower-dimensional subspace than the full parameter space suggests. Rather than discarding redundant heads, MOSAIC assigns heads to different modalities, so that each partition is sized for its own role and unnecessary cross-modal interactions are avoided by design.

\section{Methodology}
\label{sec_methodology}

Let $\mathcal{M} = \{m_1, m_2, \ldots, m_K\}$ denote a set of $K$ input modalities, and let $\mathcal{T} = \{t_1, t_2, \ldots, t_N\}$ denote a set of $N$ tasks. Each training example consists of a tuple $(\mathbf{x}^{(1)}, \mathbf{x}^{(2)}, \ldots, \mathbf{x}^{(K)}, \mathbf{y}^{(1)}, \mathbf{y}^{(2)}, \ldots, \mathbf{y}^{(N)})$, where $\mathbf{x}^{(k)}$ is the input from modality $m_k$ and $\mathbf{y}^{(n)}$ is the label for task $t_n$. The goal is to learn a shared model such that $(\hat{\mathbf{y}}^{(1)}, \ldots, \hat{\mathbf{y}}^{(N)}) = f_\theta(\mathbf{x}^{(1)}, \ldots, \mathbf{x}^{(K)})$.

\subsection{MOSAIC Architecture}

\subsubsection{Embeddings.}

Let $L$ denote the maximum sequence length processed by the shared backbone. Each modality $\mathbf{x}^{(k)}$ is encoded independently using a modality-specific feature extractor $\phi^{(k)}$, producing a sequence of $L_k$ token embeddings projected into the backbone embedding space:
\[
E^{(k)}=\mathrm{Proj}^{(k)}\!\left(\phi^{(k)}(\mathbf{x}^{(k)})\right)\in\mathbb{R}^{L_k\times d_{\mathrm{model}}}.
\]
In fusion-based architectures, modality sequences are concatenated prior to self-attention, requiring the combined sequence length to satisfy $\sum_{k=1}^{K}L_k\leq L$. In contrast, our architecture processes each modality's own sequence of $L$ tokens separately, using only that modality's dedicated subset of attention heads rather than a shared concatenated sequence, which removes this constraint and allows each modality to independently utilise the full sequence length ($L_k=L,\ \forall k$). To obtain this common length, each $E^{(k)}$ is padded or truncated to $L$ tokens, and a per-modality key padding mask prevents attention from reading padded positions. Modality sequences are matched by position index within each 30-second clip and are not explicitly time-aligned. Cross-modal correspondence is instead learned by the cross-attention layers, where each query attends over all positions of the other modality. Positional encodings are applied independently to each modality sequence to preserve intra-modal ordering before passing them to their corresponding attention-head partitions.

\begin{figure}[t]
    \centering
    \includegraphics[width=0.75\linewidth]{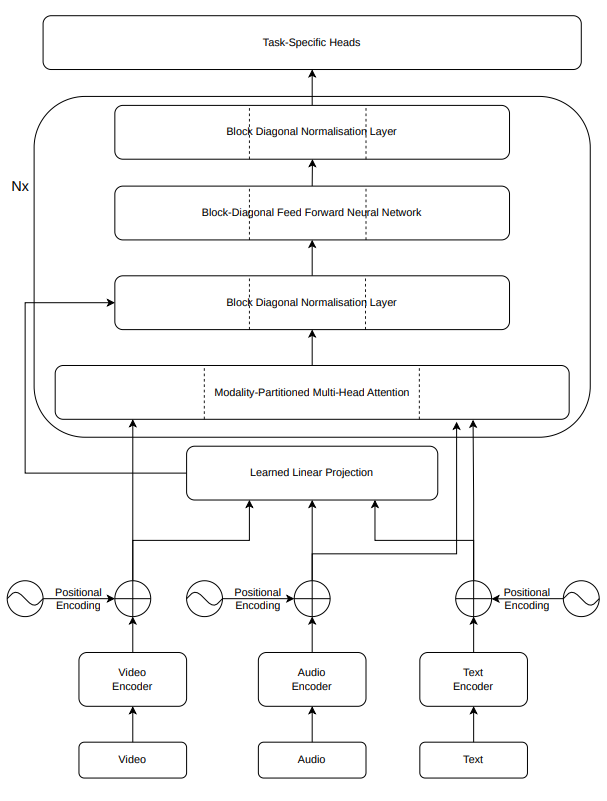}
    \caption{MOSAIC's Architecture. Dotted lines indicate separation between partitions.}
    \label{fig:mosaic_architecture}
\end{figure}

\subsubsection{Modality-Specific Attention Heads.}

Rather than allowing every attention head to operate over the same input sequence, MOSAIC partitions the attention heads of each transformer layer into $K$ disjoint modality-specific subsets. Each subset contains $H_k = H/K$ heads and is assigned to modality $m_k$.

\subsubsection{Initial Attention Layer.} For each head, the query, key, and value matrices are computed from the corresponding modality embeddings:
\[
Q_h = E^{(k)} W_h^Q, \quad
K_h = E^{(k)} W_h^K, \quad
V_h = E^{(k)} W_h^V,
\]
where $W_h^Q, W_h^K, W_h^V \in \mathbb{R}^{d_{\mathrm{model}} \times d_h}$, $d_h = d_{\mathrm{model}} / H$.
Standard scaled dot-product attention is then applied, producing head outputs $C_h \in \mathbb{R}^{L \times d_h}$, which are concatenated along the feature dimension: $C = \mathrm{Concat}(C_1, \dots, C_H) \in \mathbb{R}^{L \times d}$.

This results in a modality-partitioned representation, in which each token embedding is composed of disjoint feature subspaces corresponding to different modalities:
\[ 
z_i= \big[z_i^{(1)}\;\vert\;z_i^{(2)}\;\vert\;\cdots\;\vert\;z_i^{(K)}\big]
\]
where $d_s = d_{\mathrm{model}}/K$, and $z_i^{(k)} \in \mathbb{R}^{d_s}$ denotes the contribution of modality $m_k$ to the $i$-th token embedding.
 
Unlike conventional multimodal Transformers, which concatenate modality representations along the sequence dimension, MOSAIC concatenates modality-specific representations along the feature dimension. Consequently, every output token contains modality-specific feature partitions rather than attention over a mixed modality sequence.

\subsubsection{Residual Stream.} To preserve this modality-partitioned representation within the residual stream, each modality embedding is projected into its corresponding feature subspace via a learned linear projection,
\[
\tilde{E}^{(k)} = E^{(k)}P^{(k)},
\qquad
P^{(k)} \in \mathbb{R}^{d_{\mathrm{model}} \times d_s},
\]
yielding $\tilde{E}^{(k)} \in \mathbb{R}^{L \times d_s}$. The projected modality representations are then concatenated along the feature dimension to form the initial modality-partitioned residual stream,
$R_0 =
\mathrm{Concat}\!\left(
\tilde{E}^{(1)}, \ldots, \tilde{E}^{(K)}
\right)
\in
\mathbb{R}^{L \times d_{\mathrm{model}}}$.

All subsequent layers operate on this shared stream, but each head stays tied to its modality: it reads only that modality's slice, so self-attention never mixes modalities and cross-modal exchange occurs only in the cross-attention layers below.

\subsubsection{Block-Diagonal Feed-Forward and Normalisation Layers.}
A dense feed-forward layer would mix information across partitions and undo the separation enforced by attention. Both feed-forward weight matrices, $W_1, W_2^{\top} \in \mathbb{R}^{d_{\mathrm{model}} \times d_{\mathrm{intermediate}}}$, where $d_{\mathrm{intermediate}}$ is the intermediate dimension of the feed-forward network, are therefore parameterised as block-diagonal,
\[
W =
\begin{bmatrix}
w^{(1)} & 0 & \cdots & 0 \\
0 & w^{(2)} & \cdots & 0 \\
\vdots & \vdots & \ddots & \vdots \\
0 & 0 & \cdots & w^{(K)}
\end{bmatrix}.
\]
Each block $w^{(k)} \in \mathbb{R}^{d_s \times d_f}$, with $d_f = d_{\mathrm{intermediate}}/K$, operates only on the feature dimensions assigned to modality $m_k$. Block weights are initialised from the corresponding diagonal blocks of the pretrained feed-forward weights, while the off-diagonal entries are discarded and fixed to zero, preventing interactions between modality subspaces.

Consequently, for an input token representation $z_i$, the feed-forward layer independently transforms each modality component,
\[
\mathrm{FFN}(z)
=
W_2\,
\sigma(W_1 z + b_1)
+
b_2,
\]
\[
f_k(z^{(k)}) = \big(w_2^{(k)}\big)\sigma\!\big(z^{(k)} w_1^{(k)} + b_1^{(k)}\big) + b_2^{(k)},
\]
\[
\mathrm{FFN}(z_i)=
\big[
f_1(z_i^{(1)})
\mid
f_2(z_i^{(2)})
\mid
\cdots
\mid
f_K(z_i^{(K)})
\big],
\]
where $w_1^{(k)}$ and $w_2^{(k)}$ are the $k$-th blocks of $W_1$ and $W_2$. This preserves the modality-partitioned representation established by the attention mechanism while allowing each modality to learn independent nonlinear feature transformations.

The block structure also permits an efficient implementation: rather than multiplying by the full block-diagonal matrices, each partition is processed with its own small matrices and the outputs are concatenated along the feature dimension. This reduces the feed-forward compute and parameters by a factor of $K$ relative to a dense layer.

Layer normalisation follows the same principle. Each partition is normalised using statistics computed only over its own feature dimensions, with separate learnable scale and bias parameters, so the activation statistics of one modality cannot influence another.

\subsubsection{Cross-Attention.}
To enable inter-modal information exchange without collapsing the partitioned structure, we use cross-attention, in which the querying modality keeps its own feature partition while keys and values are drawn from another modality's partition \cite{tsai2019multimodal}.

For a head $h$ assigned to modality $m_i$, cross-attending to modality $m_j$,
\[
Q_h = z^{(i)} W_h^{Q,(i)}, \quad
K_h = z^{(j)} W_{h'}^{K,(j)}, \quad
V_h = z^{(j)} W_{h'}^{V,(j)},
\]

where $i, j \in \{1, \dots, K\}$ and $i \neq j$, and $W_h^Q, W_h^K, W_h^V$ are block-diagonal, restricted to the feature dimensions of their respective modality partitions.

Cross-attention is applied only in the later layers (4 to 12), at every odd layer, so that each modality’s representations mature within their own partitions before being exposed to the others. To ensure symmetric exchange between every pair of modalities, the $H_i$ heads assigned to $m_i$ are split into $K-1$ equal groups, each dedicated to cross-attending to one of the other modalities $m_j$. We discuss the rationale for this allocation in the Limitations section.

\paragraph{Multi-Task Classification Heads.}
Following standard multi-task learning \cite{zhang2021survey}, we attach a separate linear head $f^{(t)}(h) = W^{(t)} h + b^{(t)}$ for each task $t$ to the shared multimodal encoder, with a softmax over its $C_t$ classes.

\subsection{Modality-Specific Losses and Gradient Routing}
\label{sec:routing}

MOSAIC defines a separate loss for each modality partition, weighting every task by its overimportance and by the modality's relevance to it. Both quantities are initialised uniformly ($\alpha_t = 1/T$, $\bar{R}_{t,k} = 1/K$) and re-estimated on the validation set every $P$ epochs.

\paragraph{Task losses.}
For task $t$, with $B_t$ the number of task-$t$ samples in the batch,
\begin{equation}
\mathcal{L}_t=-\frac{1}{B_t}\sum_{b=1}^{B_t}\log p^{(t)}_{y_b}(x_b).
\end{equation}
\paragraph{Task weights.}
Following \cite{guo2018dynamic}, we track an exponential moving average (EMA) of each task's validation macro-F1,
\begin{equation}
\mathrm{F1}_t=\frac{1}{C_t}\sum_{c=1}^{C_t}\frac{2\,\mathrm{Prec}_{t,c}\,\mathrm{Rec}_{t,c}}{\mathrm{Prec}_{t,c}+\mathrm{Rec}_{t,c}},
\end{equation}
where $C_t$ is the number of classes of task $t$. With EMA coefficient $\beta_F\in[0,1)$, every $P$ epochs
\begin{equation}
\bar{F}_t\leftarrow\beta_F\,\bar{F}_t+(1-\beta_F)\,\mathrm{F1}_t,\qquad
\alpha_t=\frac{1-\bar{F}_t}{\sum_{s=1}^{T}\bigl(1-\bar{F}_s\bigr)},
\end{equation}
so that $\alpha_t\ge 0$ and $\sum_t\alpha_t=1$. Lower-performing tasks therefore receive larger weights.

\paragraph{Task--modality relevance.}
We estimate each modality's contribution to each task with Shapley values \cite{shapley1953value}, computed exactly over all modality subsets on the validation set. The value of a coalition $S\subseteq{K}$ for task $t$ is the validation macro-F1 obtained when only the modalities in $S$ are available,
\begin{equation}
v_t(S)=\mathrm{F1}_t\bigl(\text{model with modalities }{K}\setminus S\text{ ablated}\bigr),
\end{equation}
where an ablated modality has its partition of the residual stream set to zero. The Shapley value of modality $k$ for task $t$ is
\begin{equation}
\phi_{t,k}=\sum_{S\subseteq{K}\setminus\{k\}}\frac{|S|!\,(K-|S|-1)!}{K!}\Bigl[v_t\bigl(S\cup\{k\}\bigr)-v_t(S)\Bigr].
\end{equation}
These values satisfy efficiency, $\sum_{k=1}^{K}\phi_{t,k}=v_t({K})-v_t(\emptyset)$, so they decompose the gain of the full multimodal model over the fully ablated one. Because $K$ is small, the sum is evaluated exactly, requiring $2^K$ validation passes per estimation round; these are shared across all $T$ tasks, since each pass yields predictions for every task.

Negative values, which indicate that a modality hurts performance on a task, are clipped to zero and normalised across modalities,
\begin{equation}
\phi^{+}_{t,k}=\max\bigl(\phi_{t,k},0\bigr),\qquad
R_{t,k}=\frac{\phi^{+}_{t,k}}{\sum_{j=1}^{K}\phi^{+}_{t,j}},
\end{equation}
with $R_{t,k}=1/K$ if the denominator is zero. With EMA coefficient $\beta_R\in[0,1)$, the relevance is smoothed as
\begin{equation}
\bar{R}_{t,k}\leftarrow\beta_R\,\bar{R}_{t,k}+(1-\beta_R)\,R_{t,k},
\end{equation}
so that $\sum_k\bar{R}_{t,k}=1$ for every task.

\paragraph{Loss and gradient routing.}
The loss for partition $m_k$ weights each task loss by both quantities, while the shared parameters use the $\alpha$-weighted total loss:
\begin{equation}
\mathcal{L}^{(k)}=\sum_{t=1}^{T}\bar{R}_{t,k}\,\alpha_t\,\mathcal{L}_t,\qquad
\mathcal{L}_{\mathrm{total}}=\sum_{t=1}^{T}\alpha_t\,\mathcal{L}_t.
\end{equation}
The parameters are partitioned into modality-specific sets $\theta^{(k)}$ and a shared set $\theta^{(\mathrm{shared})}$. The set $\theta^{(k)}$ contains the embedding projection $P^{(k)}$, the query, key, and value projections of every head assigned to $m_k$, the attention output projection, the feed-forward weights and biases, and the layer-normalisation scale and bias parameters of partition $k$.
\subsubsection{Gradient Routing.} Each modality-specific subset $\theta^{(k)}$ is updated only with the gradient of its own loss,
\begin{equation}
\nabla_{\theta^{(k)}}\mathcal{L}^{(k)}=\sum_{t=1}^{T}\bar{R}_{t,k}\,\alpha_t\,\nabla_{\theta^{(k)}}\mathcal{L}_t,
\end{equation}
so its update is shaped both by the modality's estimated relevance to each task ($\bar{R}_{t,k}$) and by that task's importance ($\alpha_t$). The shared parameters $\theta^{(\mathrm{shared})}$ are instead updated using the gradient of the full objective, $\nabla_{\theta^{(\mathrm{shared})}} \mathcal{L}_{\mathrm{total}} = \sum_{t=1}^{T} \alpha_t\,\nabla_{\theta^{(\mathrm{shared})}} \mathcal{L}_t$, which weights tasks by $\alpha_t$ alone.

\section{Experiments}

\subsection{Experimental Setup}

\subsubsection{Datasets.}
We select the Corpus of Everyday Japanese Conversation (CEJC)~\cite{koiso2022design}, obtained via an academic licence from the National Institute for Japanese Language and Linguistics, as it is, to our knowledge, the largest naturalistic multimodal corpus jointly annotated across all four target tasks, enabling genuine multi-task evaluation rather than one assembled from disjoint single-task datasets. CEJC comprises 152 hours of multimodal conversational recordings annotated for conversational style (\textit{Style}), activity (\textit{Activity}), location (\textit{Place}), and interpersonal relationship (\textit{Relation}), segmented into 30-second clips (14,884/1,986/1,756 train/dev/test). Following~\cite{chiba2023dialogue}, rare classes are merged to alleviate class imbalance, and multi-label \textit{Relation} annotations are resolved by selecting the most distant relationship label.

\subsubsection{Baselines.}
We compare MOSAIC against unimodal models (video, audio, text) and two fusion-based baselines: AVL-BERT~\cite{chiba2023dialogue}, which uses ResNet-50 and VGGish as visual and audio encoders, and AVL-BERT (Upgraded), which swaps in MOSAIC's pretrained encoders (DINOv2, HuBERT, BERT-Japanese~\cite{oquab2023dinov2, hsu2021hubert, cl-tohoku-bert-japanese}). All models are initialised from the same BERT-Base~\cite{devlin2019bert} backbone and prediction heads. We report per-task and average macro-F1. Each configuration was trained with a fixed seed. Implementation details are in the appendix, and code will be released upon publication.

\subsubsection{Implementation Details.}
The shared backbone across all models is BERT-Base (12 layers, 12 attention heads, hidden dimension $d_{\mathrm{model}} = 768$), with a maximum sequence length of $512$. MOSAIC partitions attention heads across $K=3$ modalities (video, audio, text), assigning $H_k = 4$ heads to each modality, with a per-modality feature dimension $d_s = 256$. Modality-specific inputs are encoded using frozen pretrained encoders: DINOv2~\cite{oquab2023dinov2} for video, HuBERT~\cite{hsu2021hubert} for audio, and BERT-Japanese~\cite{cl-tohoku-bert-japanese} for text. Cross-attention layers are interleaved from layer 4 to layer 12 (inclusive).

All models are optimised using AdamW with a weight decay of $0.01$ and a batch size of $16$. Partitioned parameters, adapted from pretrained BERT-Base weights, use a lower learning rate of $1\mathrm{e}{-6}$ to preserve pretrained representations, while shared parameters, which are newly initialised, use a higher learning rate of $2\mathrm{e}{-5}$ to support faster learning from scratch.

All source code required for conducting and analysing the experiments will be made publicly available upon publication of the paper.

\subsection{Main Results}

\begin{table*}[t]
\centering
\caption{Macro-averaged F1 scores on the CEJC test set across four classification tasks. Best results are shown in \textbf{bold} and second-best results are \underline{underlined}.}
\label{tab:main_results}
\begin{tabular}{lccccc}
\toprule
\textbf{Model} & \textbf{Style} & \textbf{Place} & \textbf{Activity} & \textbf{Relationship} & \textbf{Avg.} \\
\midrule

Video (DINOv2)              & 0.562 & 0.674 & 0.289 & 0.561 & 0.522 \\
Audio (HuBERT)              & 0.620 & 0.641 & 0.262 & 0.591 & 0.529 \\
Text (BERT-Japanese)        & 0.548 & 0.303 & 0.201 & 0.558 & 0.403 \\

\midrule

AVL-BERT                    & 0.648 & 0.592 & 0.238 & 0.684 & 0.541 \\
AVL-BERT (Upgraded)         & \underline{0.692} & 0.676 & 0.251 & \underline{0.712} & 0.583 \\

\midrule

MOSAIC (Single-Task)        & 0.676 & \textbf{0.763} & \textbf{0.476} & 0.689 & \underline{0.651} \\
MOSAIC (Multi-Task)         & \textbf{0.727} & \underline{0.739} & \underline{0.441} & \textbf{0.720} & \textbf{0.657} \\

\bottomrule
\end{tabular}
\end{table*}

\begin{table*}[t]
\centering
\caption{Ablation study on CEJC. Each row replaces one component of the full MOSAIC architecture with its standard BERT-Base counterpart (see text). Macro-averaged F1 scores reported.}
\label{tab:ablation}
\begin{tabular}{lccccc}
\toprule
\textbf{Model} & \textbf{Style} & \textbf{Place} & \textbf{Activity} & \textbf{Relationship} & \textbf{Avg.} \\
\midrule
MOSAIC (full)                              & 0.727 & \textbf{0.739} & \textbf{0.441} & \textbf{0.720} & \textbf{0.657} \\
\midrule
\quad w/o Modality-Specific Attention      & 0.699 & 0.678 & 0.322 & 0.604 & 0.576 \\
\quad w/o Block-Diagonal MLP              & \textbf{0.729} & 0.710 & 0.358 & 0.664 & 0.615 \\
\quad w/o Cross-Attention Layers           & 0.687 & 0.718 & 0.341 & 0.603 & 0.587 \\
\quad w/o Modality-Specific Losses   & 0.719 & 0.708 & 0.348 & 0.628 & 0.601 \\
\bottomrule
\end{tabular}
\end{table*}

Table~\ref{tab:main_results} shows that unimodal baselines are highly uneven across tasks: audio (HuBERT) is strongest on Style, video (DINOv2) leads on Place among the unimodal models, and text (BERT-Japanese) trails on every task except Relationship, where it is broadly competitive. This confirms that no single modality is sufficient across all four tasks, motivating a fusion approach. AVL-BERT improves over the best unimodal model on Style and Relationship but underperforms the video-only baseline on both Place (0.592 vs.\ 0.674) and Activity (0.238 vs.\ 0.289), suggesting that unconstrained fusion can dilute a modality's useful signal on tasks where it is individually strong. Upgrading the fusion backbone (AVL-BERT Upgraded) recovers the video-only level on Place (0.676 vs.\ 0.674) but not on Activity (0.251 vs.\ 0.289), and yields the best baseline average (0.583). It still trails MOSAIC (Multi-Task) on every task. MOSAIC outperforms both unimodal and fusion baselines on average, with the largest gain on the hardest task, Activity, where the strongest baseline (video-only) reaches 0.289 macro-F1 against 0.441 for MOSAIC (Multi-Task) and 0.476 for MOSAIC (Single-Task), gains of 15.2 and 18.7 points. This is consistent with Activity being the task most dependent on precise cross-modal grounding, which the modality-partitioned attention and targeted cross-attention layers in MOSAIC are specifically designed to preserve. Multi-task training slightly reduces Place and Activity but improves Style and Relationship, yielding a marginally higher average (0.657 vs.\ 0.651). This suggests that gradient routing may help balance the benefits of shared learning against interference between tasks, although the small average gain and the remaining drops on Place and Activity indicate that some negative transfer persists. Overall, MOSAIC's gains stem not from access to more modalities, but from architectural control over how and when modalities interact per task.

\subsection{Ablation Study}

Table~\ref{tab:ablation} isolates the contribution of each component by replacing it in the full MOSAIC model with its standard BERT-Base counterpart, leaving all other components unchanged. Modality-specific attention is replaced by dense self-attention over the modality-partitioned residual stream, the block-diagonal MLP by the dense feed-forward network, and the cross-attention layers by standard layers. For modality-specific losses, all parameters are updated with the shared objective $\mathcal{L}_{\mathrm{total}} = \sum_t \alpha_t \mathcal{L}_t$ using the same task weights $\alpha_t$, so only the relevance matrix $R$ is removed.

Removing modality-specific attention causes the largest drop (0.657 $\rightarrow$ 0.576), especially on Activity (0.441 $\rightarrow$ 0.322) and Relationship (0.720 $\rightarrow$ 0.604), suggesting that modality-specific attention parameters are important for preserving fine-grained modality signal. Removing cross-attention layers produces the second-largest drop (0.657 $\rightarrow$ 0.587), again concentrated on Activity and Relationship, indicating that explicit inter-modal interaction is useful beyond modality-specific processing alone; Place is largely unaffected (0.739 $\rightarrow$ 0.718), suggesting it relies more on single-modality cues. Removing modality-specific losses lowers the average from 0.657 to 0.601, with Activity and Relationship again most affected, which may indicate that interference is most harmful to the hardest tasks in the absence of relevance-based routing. Replacing the block-diagonal MLP with a fully dense one causes the smallest drop (0.657 $\rightarrow$ 0.615); the dense variant has more parameters, which may partly offset the loss of modality-wise structure, and Style even improves slightly (0.727 $\rightarrow$ 0.729), consistent with it being the easiest signal for unrestricted dense mixing to exploit from the dominant modality.

\subsection{Parameter Efficiency}

\begin{table}[t]
\centering
\caption{Parameter counts and average F1 for AVL-BERT and MOSAIC variants.}
\label{tab:param_count}
\resizebox{\columnwidth}{!}{%
\begin{tabular}{lccc}
\toprule
\textbf{Model} & \textbf{Total Backbone Params} & \textbf{Trainable Params} & \textbf{Avg F1} \\
\midrule
AVL-BERT (upgraded)     & 109,482,240 & 109,482,240 & 0.583 \\
MOSAIC (no fine-tuning) & 30,813,723 & 610,587 & 0.639 \\
MOSAIC (Best)           & 30,813,723 & 11,846,400 & 0.657 \\
\bottomrule
\end{tabular}
}
\end{table}

Parameter counts exclude the frozen modality encoders (DINOv2, HuBERT and BERT-Japanese), which are identical across all systems and are not trained.

Table~\ref{tab:param_count} highlights the parameter efficiency of MOSAIC relative to AVL-BERT. The full MOSAIC backbone contains only 30.8M parameters, a 72\% reduction from AVL-BERT's 109.5M, while still achieving a higher average F1 (0.657 vs.\ 0.583). Notably, MOSAIC with no fine-tuning, using just 610K trainable parameters (0.56\% of AVL-BERT's trainable parameters), already surpasses AVL-BERT's performance (0.639 vs.\ 0.583), demonstrating that the modality-partitioned architecture provides strong inductive bias even before task-specific adaptation. Fine-tuning MOSAIC's parameters (11.8M trainable) yields a further improvement to 0.657, showing that the architecture's efficiency gains do not limit the model's ability to specialise to individual tasks. Together, these results indicate that MOSAIC's structured partitioning enables competitive or superior performance with substantially fewer parameters, both in total and in what must be trained.

\section{Conclusion}
We introduced MOSAIC, a modality-partitioned transformer for multimodal multi-task learning that addresses task-modality interference through structured representation separation and gradient routing. By preserving modality-specific feature spaces while enabling controlled cross-modal interaction, MOSAIC improves task specialisation without separate modality-specific transformer stacks. Experiments on CEJC show improved multi-task performance with substantially fewer backbone parameters than fusion-based alternatives, alongside reduced task interference supported by gradient analysis. These results highlight the value of explicitly modelling task-modality dependencies in multimodal architecture design.

\section*{Limitations and Future Work}

Our evaluation is restricted to a single Japanese dataset, CEJC, with three modalities (audio, video, and text), a BERT-Base backbone, and modalities matched by position index within each clip rather than explicitly time-aligned. We therefore do not know how well the findings transfer to other multimodal multi-task benchmarks, languages, label sets, or tasks whose modality dependencies differ from the four studied here. Evaluation on additional datasets, a broader range of modalities, explicit temporal alignment, and larger pretrained backbones remains an important direction. We also compare MOSAIC against a fusion-based baseline but not against methods that address task conflict through loss weighting, gradient surgery (e.g., PCGrad), or task grouping, which we discuss in the introduction. It thus remains unclear whether these optimisation methods would narrow the gap when applied to a fusion-based model, or whether they are complementary to partitioning when combined with MOSAIC.

MOSAIC’s design also rests on several simplifying assumptions. Capacity, feature dimensions, and attention heads are allocated equally across modalities regardless of their complexity, and head assignment, cross-modal pairing, and the placement and frequency of cross-attention layers are fixed rather than learned. In particular, the $H_k$ heads of each modality are split equally among the other $K-1$ modalities. We adopt this symmetric scheme because it requires no prior assumption about which modality should attend to which, and it divides evenly in our configuration ($H_k = 4$, $K = 3$, giving two heads per modality pair). The trade-off is that every pair receives the same cross-attention capacity regardless of how useful the exchange is for the tasks at hand, so a pair that matters little for a given task consumes heads that a more informative pair could use. We did not compare alternative allocation schemes, such as head counts proportional to the relevance matrix $R$ or learned pair weights. Adaptive capacity allocation, data-driven head selection, and a learned cross-attention schedule are natural extensions.

Finally, modality-wise partitioning may limit scalability as the number of modalities grows, since each added modality brings its own parameter partition and more cross-attention pairs, while the capacity available per pair shrinks. This motivates dynamic or hierarchical partitioning schemes.

\section*{Ethical Considerations}
Our primary benchmark, the Corpus of Everyday Japanese Conversation (CEJC), was obtained by purchasing an academic licence from the National Institute for Japanese Language and Linguistics (NINJAL), and we use it solely for academic research under the terms of that licence.

\section*{Acknowledgements}
We used Claude (Anthropic) and ChatGPT (OpenAI) to edit parts of the text and to write parts of the code. All outputs were reviewed and verified by the authors, who take full responsibility for the content.

\bibliography{aaai2027}


\end{document}